\documentclass[letterpaper]{article} 
\usepackage{aaai2026}  
\usepackage{times}  
\usepackage{helvet}  
\usepackage{courier}  
\usepackage[hyphens]{url}  
\usepackage{graphicx} 
\usepackage{natbib}  
\usepackage{caption} 

\usepackage{algorithm}
\usepackage{algorithmic}

\usepackage{newfloat}
\usepackage{listings}
\DeclareCaptionStyle{ruled}{labelfont=normalfont,labelsep=colon,strut=off} 
\floatstyle{ruled}
\newfloat{listing}{tb}{lst}{}
\floatname{listing}{Listing}
\usepackage{latexsym}
\usepackage{amssymb}
\usepackage{amsmath}
\usepackage{amsthm}
\usepackage{booktabs}
\usepackage{enumitem}
\usepackage{graphicx}
\usepackage{booktabs} 
\usepackage{tabularx} 
\usepackage{array} 
\usepackage{float}
\usepackage{multirow}
\usepackage{bbding}

\title{Not Just What's There: Enabling CLIP to Comprehend Negated Visual Descriptions Without Fine-tuning}

\author{
    Junhao Xiao\textsuperscript{\rm 1},
    Zhiyu Wu\textsuperscript{\rm 2},
    Hao Lin\textsuperscript{\rm 3},
    Yi Chen\textsuperscript{\rm 1}\footnote{Corresponding authors.},
    Yahui Liu\textsuperscript{\rm 4}\footnotemark[1],\\ 
    Xiaoran Zhao\textsuperscript{\rm 5},
    Zixu Wang\textsuperscript{\rm 5},
    Zejiang He\textsuperscript{\rm 5}
}

\affiliations{
    \textsuperscript{\rm 1}Central China Normal University,  \\
    \textsuperscript{\rm 2}Fudan University, 
    \\
    \textsuperscript{\rm 3}Huazhong University of Science and Technology,
    \\
    \textsuperscript{\rm 4}Kuaishou Technology, 
    \\
    \textsuperscript{\rm 5}National University of Defense Technology
    \\
    chenyi30@mail.ccnu.edu.cn, yahui.cvrs@gmail.com
}

\usepackage{bibentry}

\begin{document}

\maketitle

\begin{abstract}
Vision-Language Models (VLMs) like CLIP struggle to understand negation, often embedding affirmatives and negatives similarly (\textit{e.g.}, matching ``no dog'' with dog images). Existing methods refine negation understanding via fine-tuning CLIP’s text encoder, risking overfitting. In this work, we propose \textsc{CLIPGlasses}, a plug-and-play framework that enhances CLIP’s ability to comprehend negated visual descriptions.
\textsc{CLIPGlasses} adapts a dual-stage design: a \textit{Lens} module disentangles negated semantics from text embeddings, and a \textit{Frame} module predicts context-aware repulsion strength, which is integrated into the modified similarity computation to penalize alignment with negated semantics, thereby reducing false positive matches.
Experiments show that CLIP equipped with \textsc{CLIPGlasses} achieves competitive in-domain performance and outperforms state-of-the-art methods in cross-domain generalization. Its superiority is especially evident under low-resource conditions, indicating stronger robustness across domains.
\end{abstract}

\begin{links}
   \link{Code}{https://github.com/Codecode-X/CLIPGlasses.git}
\end{links}

\section{Introduction}
\label{sec:introduction}

Recent advances in large-scale pretraining have advanced Vision-Language Models (VLMs), with CLIP~\cite{radford2021learning} emerging as a foundational model. CLIP enables cross-modal alignment by projecting images and texts into a shared embedding space, underpinning core tasks such as retrieval, captioning, and text-conditioned generation. 

\begin{figure}[t]
    \centering
    \includegraphics[width=\columnwidth]{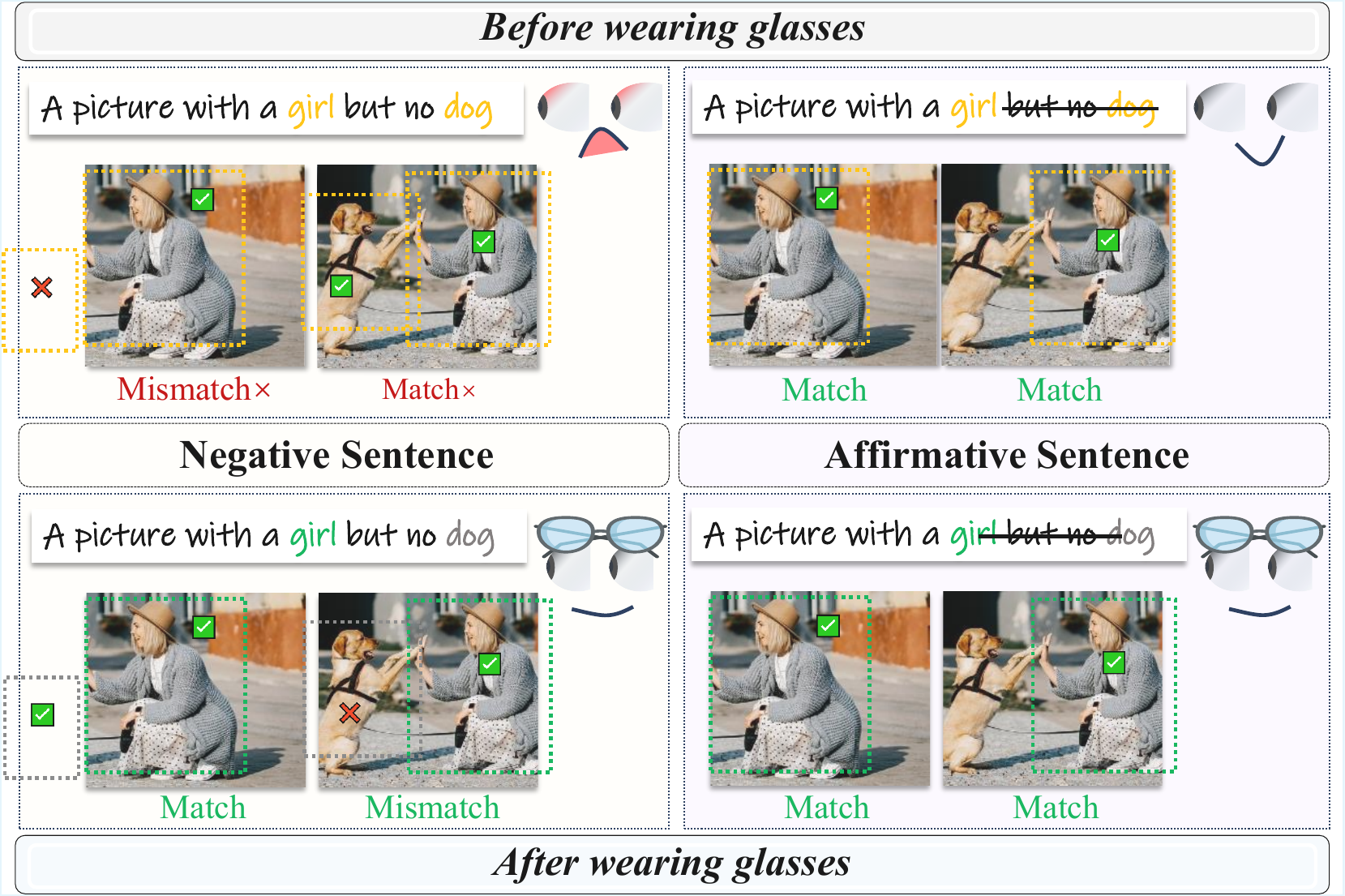}
    \caption{
        \textsc{CLIPGlasses} enhances CLIP’s capacity for negation understanding by introducing a dynamic repulsion mechanism that suppresses image-text similarity for negated concepts, thus enabling inverse matching while preserving alignment with affirmed content.
    }
    \label{fig:eyes}
\end{figure}

However, CLIP exhibits notable limitations in modeling negation semantics. 
As shown in Figure~\ref{fig:eyes}, CLIP fails to properly handle negation cues like ``\emph{no}'' or ``\emph{without}'' in text inputs, incorrectly matching negated concepts with corresponding visual content rather than recognizing their absence. This failure stems from the sparsity of negation expressions in pretraining corpora (\textit{i.e.}, only 0.7\%~\cite{NegationCLIP}), which prevents contrastive learning from effectively capturing semantic polarity reversals.

Several existing approaches have attempted to adapt CLIP models to negation-sensitive tasks through parameter fine-tuning~\cite{NegClip,alhamoud2025vision,NegationCLIP,conclip}, but these approaches poses two critical drawbacks.
First, constructing large-scale negation-annotated datasets is time-consuming and resource-intensive. Second, fine-tuning introduces the risk of catastrophic forgetting, whereby enhanced negation understanding comes at the expense of deteriorating general-purpose performance. This represents a fundamental trade-off between specialized capability and broad applicability.

To address these limitations, we draw inspiration from two key observations. First, although affirmative and negative semantics lie close in CLIP’s feature space, visualization analysis (Figure~\ref{fig:tsne}) reveals structured separability enabled by layer-specific encoding~\cite{how2024acl}, indicating the feasibility of disentangling negation-related information from CLIP-encoded text embeddings. Second, cognitive studies show that humans process negation in two stages, first identifying the negated concept, then inverting its meaning~\cite{kaup2007experiential,orenes2014negation,zuanazzi2024negation}. Guided by these insights, we design a plug-and-play framework \textsc{CLIPGlasses} that leverages CLIP’s latent negation representations and emulates this two-stage process. 

Specifically, \textsc{CLIPGlasses} extends CLIP through two lightweight modules. The \textit{Lens} module disentangles negated semantics from text embeddings, while the \textit{Frame} module predicts context-dependent repulsion strength. These components jointly enable a modified similarity computation: negated content identified by \textit{Lens} is explicitly penalized through repulsion vectors whose magnitude is dynamically scaled by \textit{Frame}'s strength predictions. This integrated process is visually summarized in Figure~\ref{fig:overview}. To coordinate these components, we employ a progressive three-stage regimen with frozen CLIP parameters: first training \textit{Lens} to disentangle negated text representations, then training \textit{Frame} to model dynamic repulsion using ground-truth negation features, and finally joint optimization of both modules with \textit{Lens} outputs to enhance synergy. In this way, instead of modifying the model’s \emph{“eyes”}, we simply let it wear \emph{“glasses”} to better perceive the negation. 

Experimental results establish \textsc{CLIPGlasses}'s balanced advantages through systematic comparisons. While fine-tuning baselines overfit negation datasets for marginal in-domain gains (CoN-CLIP: 99.70\% vs. our 96.56\% on CC-Neg-val), this incurs unwilling costs: degraded cross-domain generalization (25.70\% vs. our 34.51\% on Neg-COCO-MCQ). Under low-resource constraints (5K images), these limitations intensify—our approach surpasses CoN-CLIP by 27.45 points on CC-Neg-val and 5.29 points on Neg-COCO-MCQ. Moreover, unlike fine-tuning approaches that impair CLIP’s native zero-shot performance on standard non-negation benchmarks, our non-invasive architecture effectively preserves these intrinsic capabilities. These results substantiate the efficacy of \textsc{CLIPGlasses} in achieving transferable, semantically grounded negation understanding without sacrificing model robustness or generalization.
We summarize our key contributions as follows:
\begin{itemize}
    \item We present \textsc{CLIPGlasses}, a non-intrusive framework enhancing CLIP's negation modeling via human-inspired two-stage processing without parameter modification.
    \item We design a novel architecture, including a syntax-semantic \textit{Lens} for disentangling negation semantics, and a \textit{Frame} for modeling context-aware repulsion, and a modified similarity computation that explicitly reverses alignment with negated content.
    \item Our method attains state-of-the-art trade-offs between in-domain accuracy and cross-domain generalization, without compromising CLIP’s native zero-shot abilities.
\end{itemize}

\section{Related Work} \label{sec:related_work}

\noindent
\textbf{Vision-Language Models and Their Limitations.}
Vision-language models like CLIP~\cite{radford2021learning} demonstrate strong performance on broad cross-modal tasks, including retrieval~\cite{qin2025clip,caffagni2025recurrence,fang2024fewer,zhang2025cwnet,li2025stitchfusion,zhang2025spjfnet,li2025exploring}, captioning~\cite{kim2025vipcap,nam2025extract,li2025madakv} and VQA~\cite{xing2024clipvqa,sun2024alpha,jiang2025fedcfa} and generation~\cite{zhang2025semtalk,feng2025unified,stablediffusion,zhang2025beyond,ijcai2025p109,zhang2025echomask}, leveraging representations learned from large-scale noisy image-text pairs. 
However, they exhibit critical deficiencies in fine-grained semantic understanding, manifesting in four key areas: over-reliance on shallow statistical cues~\cite{zhang2023biomedclip,geirhos2020shortcut}, failures in compositional reasoning~\cite{NegClip,thrush2022winoground}, weak attribute grounding, and bag-of-words-like text processing that ignores syntactic structure~\cite{bagwords2025}. 
Negation understanding poses a particularly challenging case, with VLMs frequently failing to distinguish negated and affirmative inputs~\cite{morante2021recent,ma2023crepe} due to affirmation bias and scarcity of negation in pretraining data (occurring in $\leq$0.7\% of captions~\cite{NegationCLIP}), significantly impairing negation-sensitive applications, such as medical or clinical scenarios~\cite{Ko_2025_CVPR,su2025large}.

\noindent
\textbf{Prior Work on Negation Understanding.}
Existing methods universally fine-tune CLIP’s text encoder and fall into two categories based on data strategy: (1) Compositional Perturbation, typified by NegCLIP~\cite{NegClip}, which uses word-order shuffling to simulate syntactic variation in negation; and (2) Paraphrastic Augmentation, including CoN-CLIP~\cite{conclip}, NegationCLIP~\cite{NegationCLIP}, and NegBench~\cite{alhamoud2025vision}, which leverage LLM-generated negatives to enrich contrastive training. 
While effective in-domain, these methods overwrite CLIP’s pretrained representations, risking overfitting, catastrophic forgetting~\cite{zhai2024investigating,jha2024clap4clip,li2024revisiting}, and poor cross-domain generalization due to the lack of explicit negation modeling. In contrast, \textsc{CLIPGlasses} introduces dedicated negation reasoning via architectural extensions, preserving CLIP’s zero-shot capabilities without modifying its parameters.

\section{Preliminary: Two-Stage Modeling for Negation Understanding} \label{sec:preliminary}

\begin{figure}[t]
    \centering
    \includegraphics[width=\columnwidth]{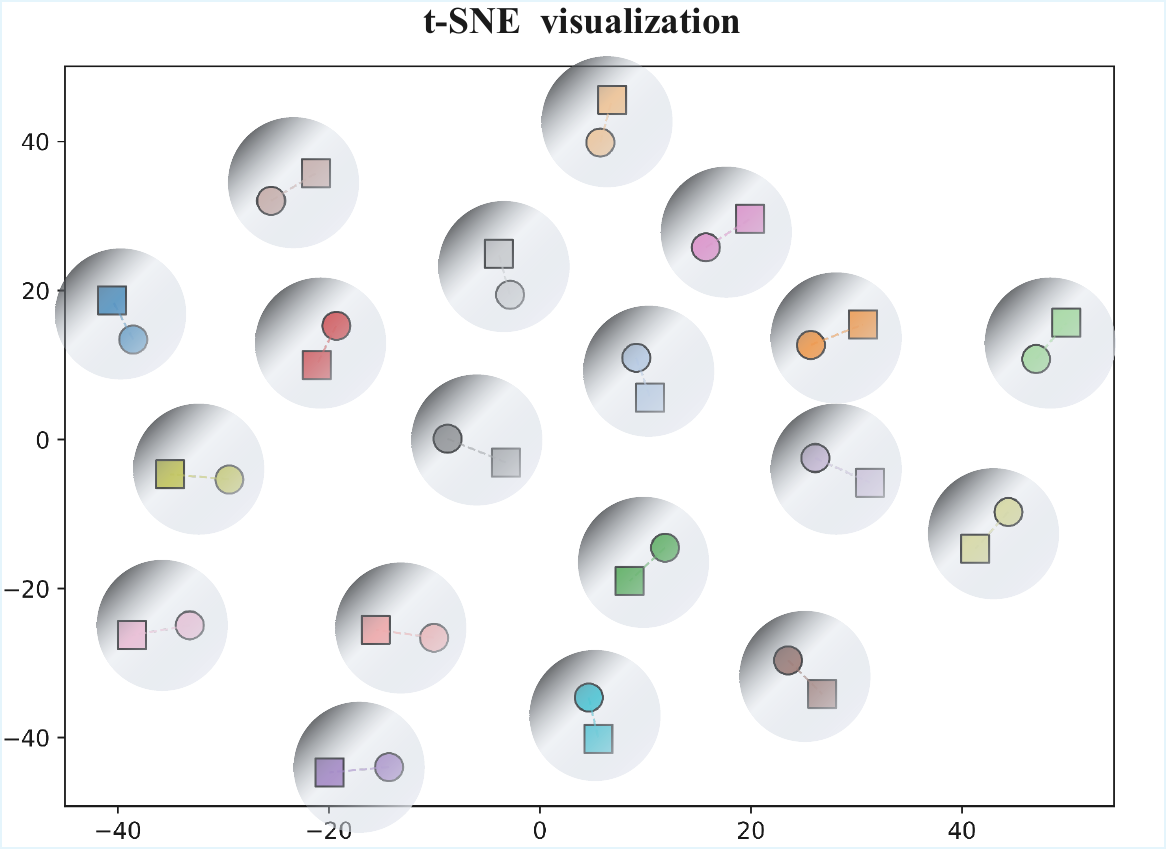}
    \caption{
        t-SNE~\cite{tsne} visualization of CLIP text features for multiple positive-negative sentence pairs (\textit{e.g.}, ``\emph{there is a woman}'' vs. ``\emph{there is not a woman}''). Circles and squares denote positive and negative forms, colors distinguish different pairs. Feature clusters across pairs are well-separated, showing CLIP’s strong instance-level discrimination. However, positive and negative features within individual pair remain closely positioned, indicating that while CLIP has limited negation modeling capabilities, there exists clear potential for semantic disentanglement.
    }
    \label{fig:tsne}
\end{figure}

Although affirmative and negative semantics are often embedded in close proximity within CLIP’s feature space, visualization analysis (See Figure~\ref{fig:tsne}) reveals a promising potential for semantic disentanglement. This separability arises from layer-specific negation encoding mechanisms~\cite{how2024acl} and forms a solid basis for targeted extraction of negation-related information.

Additionally, findings from cognitive neuroscience indicate that humans typically process negation in two distinct stages~\cite{kaup2007experiential,orenes2014negation,zuanazzi2024negation}: First, by identifying the target object or concept being negated; Second, by inverting its semantic implication to derive the negated meaning. Inspired by this two-stage cognitive mechanism, we propose a corresponding computational modeling strategy to better handle negation in vision-language alignment. In the first stage, the model explicitly identifies and extracts negation semantics from textual features. In the second stage, it modulates the image-text similarity by explicitly suppressing alignment with negated concepts. 

\begin{figure*}[t]
    \centering
    \includegraphics[width=\textwidth]{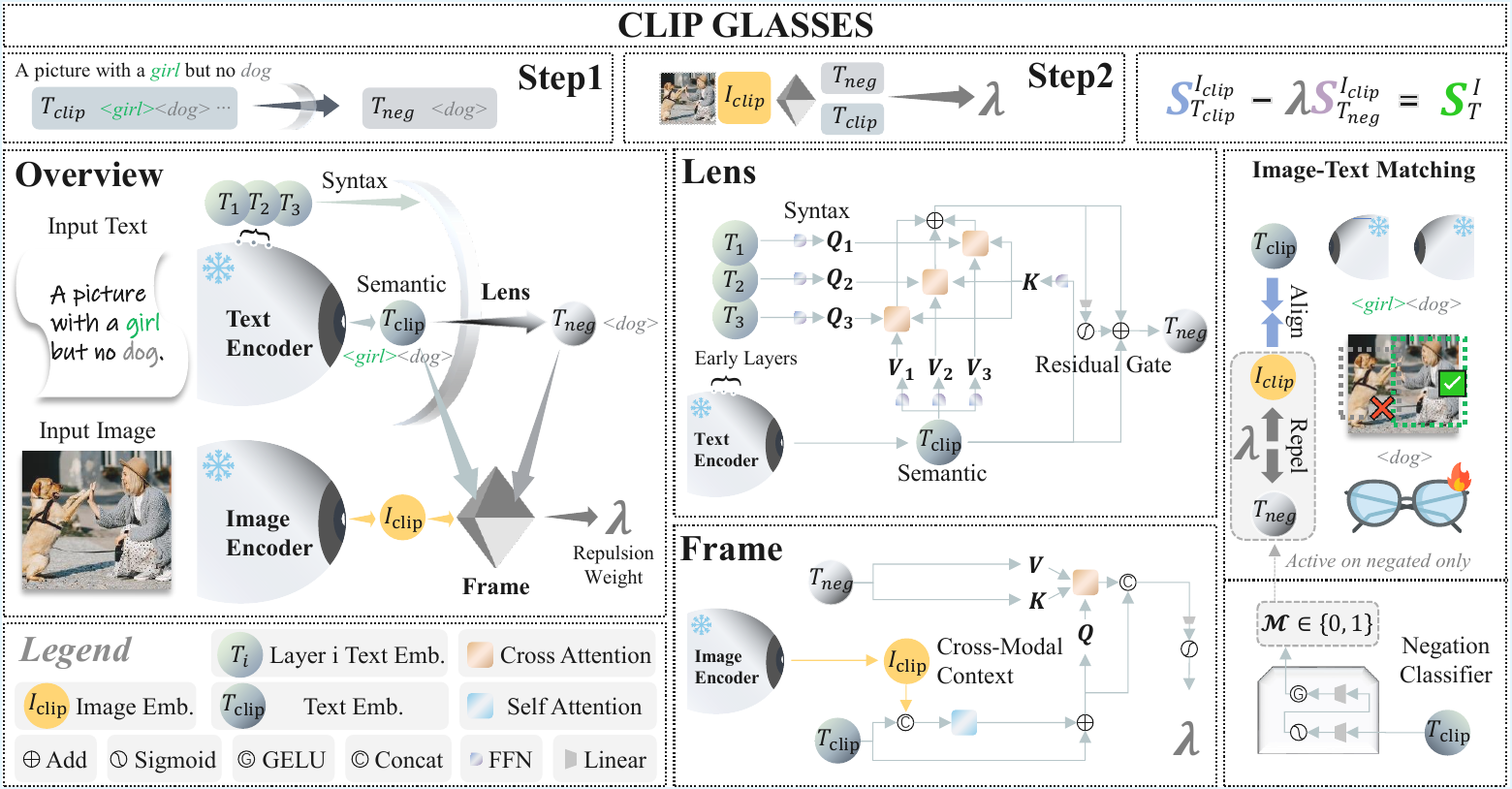}
    \caption{\textsc{CLIPGlasses} enhances CLIP's capability to model negative semantics by introducing two modules: \textit{Lens} and \textit{Frame}. \textit{Lens} disentangles negated concepts (\textit{e.g.}, ``\emph{dog}'' in ``no \emph{dog}'') from the text embedding $T_{\text{clip}}$. \textit{}{Frame} dynamically predicts a repulsion strength $\lambda$ based on cross-modal context. The final similarity score is computed as $S = S_{\text{I2T}} - \lambda \cdot S_{\text{I2T}}^{\text{neg}}$, aligning images with affirmed content while repelling from negated concepts when negation is present in the text.}
    \label{fig:overview}
\end{figure*}

\section{Methodology}

To instantiate the two-stage modeling paradigm outlined above, we propose \textsc{CLIPGlasses}, a modular extension to CLIP for negation-aware vision-language alignment. It comprises two components: \textit{Lens}, which disentangles negated semantics from text embeddings, and \textit{Frame}, which predicts context-dependent repulsion strength. These components are integrated into a modified similarity computation that penalizes alignment with negated content. An overview of the architecture is shown in Figure~\ref{fig:overview}. 
We describe the architecture and functionality of each module in the following and then present the final matching strategy.

\subsection{Lens: Syntax–Semantic Dual-Stream Architecture}

While CLIP’s text encoder effectively models affirmative semantics, it lacks explicit mechanisms for capturing the structural and contextual characteristics of negation. As a result, its output features often fail to distinguish the semantic differences inherent in negated expressions. To address this limitation, we propose the Lens module, a syntax-semantic dual-stream architecture that explicitly models negation-aware text representations via hierarchical attention.

\noindent\textbf{Syntactic Stream.}  
Natural language negation frequently manifests through syntactic patterns (e.g., auxiliary constructions like ``\emph{do not}'', adverbials like ``\emph{never}'') that exhibit local structural dependencies. To capture these cues, we extract features ${T_1, T_2, T_3}$ from the first three layers of the CLIP text encoder, which encode low-level syntactic information~\citep{how2024acl}. Each layer's features are projected into a shared latent space via the transformation:
\begin{equation}
\mathcal{P}(X) = \mathrm{LN}(\mathrm{GELU}(W X))
\end{equation}
where $W$ denotes a learnable weight matrix unique to each projection instance, $\mathrm{GELU}$ is the Gaussian Error Linear Unit~\cite{GELU}, and $\mathrm{LN}$ represents Layer Normalization~\cite{LN}. The resulting query vectors are:
\begin{equation}
Q_i = \mathcal{P}_{q,i}(T_i), \quad i\in\{1,2,3\}
\end{equation}

\noindent\textbf{Semantic Stream.}  
While syntactic structures signal negation expression, precise interpretation requires global semantic comprehension. For instance, in the sentence ``\emph{He claimed he finished the work, but actually hadn't}'', the negation scope of ``\emph{hadn't}'' depends critically on the antecedent context (``\emph{finished the work}'') and the contrastive conjunction (``\emph{but}''). To capture such context-dependent negation, we utilize the final-layer output $T_{\mathrm{clip}}$ of the CLIP text encoder~\cite{lin2024tagclip,jing2024fineclip}, which provides globally contextualized representations for generating:
\begin{equation}
K = \mathcal{P}_{k}(T_{\mathrm{clip}}), \quad V_i = \mathcal{P}_{v,i}(T_{\mathrm{clip}}), \quad i\in\{1,2,3\}
\end{equation}

\noindent\textbf{Hierarchical Attention Fusion.}  
Syntactic representations from different layers encode linguistic features at varying granularities: $Q_1$ captures local token interactions (e.g., negation particles and adjacent verbs), while the deeper $Q_3$ encodes phrasal-level dependencies. To dynamically integrate these hierarchical patterns with global semantics, we compute attention weights between each $Q_i$ and the semantic key $K$, modulated by learnable scalars $\alpha_i$ that adaptively calibrate each syntactic level's contribution:
\begin{equation}
T_{\mathrm{attn}} = \sum_{i=1}^3 \mathrm{softmax}\left(\frac{Q_i K^\top}{\sqrt{D}} + \log \alpha_i \right) V_i
\end{equation}

\noindent\textbf{Residual Gating.}  
While hierarchical attention enriches the representation with negation-relevant structure, over-reliance on it may cause semantic drift or loss of core content. To address this,we apply a residual gate that adaptively blends the attended representation with the original CLIP features:
\begin{equation}
g = \sigma(W_g T_{\mathrm{attn}} + b_g)
\end{equation}
\begin{equation}
    T_{\mathrm{neg}} = \mathrm{FFN}\bigl(g \odot T_{\mathrm{attn}} + (1 - g) \odot T_{\mathrm{clip}}\bigr)
\end{equation}
where \(\sigma\) is the sigmoid and \(\odot\) denotes Hadamard multiplication. The gate \(g\) acts as a soft selector, allowing the model to amplify structural adjustments only when necessary. The bias \(b_g\) is initialized to a negative value to favor original features during early training. Finally, a feed-forward neural network (FFN) refines the output representation.

\subsection{Frame: Cross-Modal Dynamic Repulsion Weight Generator}
Negation in language varies in intensity and contextual nuance (\textit{e.g.}, ``\emph{not}'' vs. ``\emph{may not}''), directly affecting the degree to which negated concepts influence vision-language alignment. To address this, the Frame module dynamically estimates a repulsion weight $\lambda$ based on joint image-text representations, which governs the strength of semantic inversion.

\noindent\textbf{Cross-Modal Context.}
Negation is inherently linguistic, yet its interpretation is often grounded in visual context~\cite{sun2021rpbert,janssens2024integrating}. To estimate $\lambda$ in a context-aware manner, we first encode cross-modal interactions by allowing each modality to attend to the other.

Before fusing the textual and visual features, we apply L2 normalization to eliminate scale discrepancies that could hinder effective attention. For numerical stability, a small constant \(\epsilon\) is added in the denominator when normalizing the negated text feature:
\begin{equation}
\hat{I}_{\mathrm{clip}} = \frac{I_{\mathrm{clip}}}{\|I_{\mathrm{clip}}\|_2}, \quad 
\hat{T}_{\mathrm{clip}} = \frac{T_{\mathrm{clip}}}{\|T_{\mathrm{clip}}\|_2}, \quad 
\hat{T}_{\mathrm{neg}} = \frac{T_{\mathrm{neg}}}{\|T_{\mathrm{neg}}\|_2 + \epsilon}
\end{equation}

To capture bidirectional cross-modal dependencies, we employ a joint self-attention mechanism~\cite{attention} over concatenated text and image features. Compared to directional cross-attention, this symmetric early fusion enables both modalities to attend to each other jointly, avoiding representational bias. Formally:
\begin{equation}
\begin{aligned}
    F_{\mathrm{T\&I}} &= \operatorname{SelfAttn}\left([\hat{T}_{\mathrm{clip}}; \hat{I}_{\mathrm{clip}}]\right) \in \mathbb{R}^{2 \times D}
\end{aligned}
\end{equation}

Since \(\lambda\) modulates textual semantics, its estimation should be grounded in a representation that is text-centric yet contextually enriched by visual cues. We thus retain only the text-side output $F_{\mathrm{T\&I}}^{(1)}$ from the joint self-attention. To preserve original semantics while enabling adaptive fusion, we apply a residual connection:
\begin{equation}
\begin{aligned}
T_{\mathrm{fuse}} &= \alpha\, \hat{T}_{\mathrm{clip}} + F_{\mathrm{T\&I}}^{(1)}
\end{aligned}
\end{equation}
where \(\alpha \in [0, 1]\) is a learnable scalar (initialized to 0.1) that controls the balance between the original and the context-enhanced text features.

\noindent\textbf{Dynamic Repulsion Weight.}
Estimating the repulsion strength $\lambda$ hinges on highlighting negated semantics relevant to the fused text-visual context. Since negation inherently involves contrasting the original semantics with negated concepts, a mechanism that attends to the negated features conditioned on the enriched textual context is essential.

To this end, we first compute a cross-attention output \(C\), where the fused representation \(T_{\mathrm{fuse}}\) acts as the query, and the negated semantics \(\hat{T}_{\mathrm{neg}}\) serve as keys and values. This allows the model to dynamically align and weight the negated information relevant to the current context:
\begin{equation}
C = \operatorname{CrossAttn}(
\underbrace{T_{\mathrm{fuse}}}_{Q},\;
\underbrace{\hat{T}_{\mathrm{neg}}}_{K},\;
\underbrace{\hat{T}_{\mathrm{neg}}}_{V})
\end{equation}

Next, the repulsion weight \(\lambda\) is estimated by projecting the concatenation of the fused feature \(T_{\mathrm{fuse}}\) and the cross-attention output \(C\) through a learnable linear layer followed by a sigmoid activation:
\begin{equation}
\lambda = \sigma \left( W_\lambda [T_{\mathrm{fuse}} ; C] + b_\lambda \right)
\end{equation}

\subsection{Image-Text Matching}

With the negated text representation \(T_{\mathrm{neg}}\) generated by the \textit{Lens} module and the context-sensitive repulsion weight \(\lambda\) estimated by the \textit{Frame} module, we now describe how these components jointly contribute to the final image-text matching objective. The core idea is to preserve the original alignment ability of CLIP while penalizing false positives caused by visual-textual agreement with negated semantics. 

The final similarity score thus combines a standard matching term with a negation-aware repulsion component. For generality across both text-to-image and image-to-text retrieval, we adopt a unified notation where \(q\) denotes the query modality (either text or image) and \(k\) denotes the key modality (the corresponding paired instance).

\noindent\textbf{Basic Similarity Calculation.}  
We begin with the original CLIP similarity as the base score. This term reflects the semantic alignment between the image and the text:
\begin{equation}\label{eq:score_base}
    S_{\text{base}}(q, k) = \exp(\theta_T) \cdot \frac{q^\top k}{\|q\|_2 \|k\|_2}
\end{equation}
where \(\theta_T\) is the temperature parameter of the pretrained CLIP model.

\noindent\textbf{Negation-aware Similarity Adjustment.}  
While the base similarity captures standard semantic alignment, it is insufficient for handling negation. If the text explicitly negates certain visual content, the model must reduce the matching score accordingly. To this end, we introduce a negation-aware repulsion term that reflects how well the image aligns with the negated semantics. The repulsion term is computed by:
\begin{equation}\label{eq:repulsion}
    \mathcal{R}_{\mathrm{neg}} = \lambda \cdot \max\left( \exp(\theta_T) \cdot \frac{q^\top k_{\mathrm{neg}}}{\|q\|_2 \|k_{\mathrm{neg}}\|_2},\ 0 \right)
\end{equation}
where the \(\max(\cdot, 0)\) operation ensures that the repulsion term only penalizes positive alignment with negated concepts, preventing undesirable score inflation.

\noindent\textbf{Final Similarity Score.}  
To avoid unnecessary interference in affirmative cases and better preserve CLIP’s performance on general (non-negated) tasks, we further introduce a conditional mask that selectively activates the repulsion term based on the presence of negation.

Concretely, we define a binary decision variable \(\mathcal{M} \in \{0,1\}\) that determines whether the similarity score should be corrected. The decision is made by a lightweight negation classifier \(\mathcal{G}: \mathbb{R}^D \rightarrow [0,1]\), which predicts the probability of negation from the original text representation:

\begin{eqnarray}\label{eq:mask}
    \mathcal{M} = \mathbb{I}[\mathcal{G}(T_{clip}) > \tau_{neg}]
\end{eqnarray}
where \(\tau_{\mathrm{neg}}\) is a threshold (default 0.5). The classifier \(\mathcal{G}\) is implemented as a two-layer MLP.

The repulsion term \(\mathcal{R}_{\mathrm{neg}}\) is applied only when \(\mathcal{M} = 1\), ensuring that the correction is restricted to negated inputs while maintaining standard CLIP behavior elsewhere. 

The final similarity score integrates the base CLIP similarity with the negation-aware repulsion term under the control of the conditional mask:
\begin{equation}\label{eq:final}
    S = S_{\mathrm{base}} - \mathcal{M} \cdot \mathcal{R}_{\mathrm{neg}}
\end{equation}

To ensure stable optimization, the mask is fixed to $\mathcal{M} = 1$
during training, allowing gradients to propagate consistently.
It is only activated at inference time to selectively suppress
negated alignments.

\subsection{Training Strategy}
We adopt a staged training strategy to progressively model cross-modal negation semantics. The process consists of three phases: (1) training the Lens module to capture fine-grained representations of negated text; (2) training the Frame module to establish dynamic interactions between negation and image features; and (3) joint optimization to enhance collaboration between the two modules.

\begin{table*}[ht]
   \setlength{\tabcolsep}{4pt}
  
  \newcolumntype{L}[1]{>{\raggedright\arraybackslash}p{#1}}
  \newcolumntype{C}[1]{>{\centering\arraybackslash}p{#1}}

   \begin{tabularx}{0.98\textwidth}{
    C{0.12\textwidth}
    C{0.12\textwidth}
    C{0.12\textwidth}
    C{0.12\textwidth}
    C{0.11\textwidth}
    C{0.15\textwidth}
    C{0.10\textwidth}
  }

    \toprule
    \multirow{3}{*}{\textbf{Method}}
    & \multirow{3}{*}{\textbf{Training Set}}
    & \textbf{Training Size}
    & \multicolumn{4}{c}{\textbf{Testing}} 
    \\

    \cmidrule(r){4-7} 
    & 
    & \multicolumn{1}{c}{\textbf{(Images/Captions)}}
    & \textbf{Testing Set }
    & \textbf{Accuracy}
    & \textbf{Testing Set}
    & \textbf{Accuracy}
    \\

    \midrule
    NegCLIP & COCO-caption & 82K/31K & CC-Neg-val & 62.63\% & Neg-COCO-MCQ & 10.20\%
    \\
    CoN-CLIP & CC-Neg & 188K/376K & CC-Neg-val & \textbf{99.70\%} & Neg-COCO-MCQ & 25.70\%
    \\
    CLIPGlasses & CC-Neg & 188K/376K & CC-Neg-val & 96.56\%\textsuperscript{-3.14\%} & Neg-COCO-MCQ & \textbf{34.51\%}\textsuperscript{+8.81\%}
    \\
    \midrule
    CoN-CLIP & Neg-COCO-R & \textbf{5K/22K} & CC-Neg-val & 65.91\% & Neg-COCO-MCQ & 30.61\%
    \\    
    CLIPGlasses & Neg-COCO-R & \textbf{5K/22K} & CC-Neg-val & \textbf{93.36\%}\textsuperscript{+27.45\%} & Neg-COCO-MCQ & \textbf{35.90\%}\textsuperscript{+5.29\%}
    \\
    
    \bottomrule
  \end{tabularx}
  \caption{Performance comparison highlighting the trade-off between in-domain fitting and cross-domain generalization.}
  \label{tab:compare_tbl}
\end{table*}

\noindent\textbf{Stage 1: Independent Training of the Lens Model.}
In this stage, we extract the negated object from the input text and generate a short prompt (\textit{e.g.}, ``\emph{This image shows a \{negobj\}}''). This prompt is then passed to the CLIP text encoder to extract the ground truth negation feature $T_{\mathrm{obj}}^{\mathrm{neg}} \in \mathbb{R}^D$ as the supervision signal.

We design a training objective that combines the semantic similarity loss $\mathcal{L}_{\mathrm{sim}}$ and the cross-modal alignment loss $\mathcal{L}_{\mathrm{align}}$ to ensure the learned negation features are both accurate and aligned with image features:
\begin{equation}
    \mathcal{L} = \mathcal{L}_{\mathrm{sim}} + \delta \mathcal{L}_{\mathrm{align}}
\end{equation}

The semantic similarity loss $\mathcal{L}_{\mathrm{sim}}$ ensures semantic consistency between the predicted negation feature $T_{\mathrm{neg}}$ and the ground truth negation feature $T_{\mathrm{obj}}^{\mathrm{neg}}$:
\begin{equation}
    \mathcal{L}_{\mathrm{sim}} = 1 - \frac{1}{B} \sum_{j=1}^{B} \frac{T_{\mathrm{neg}}^{(j)} \cdot T_{\mathrm{obj}}^{\mathrm{neg},(j)}}{\|T_{\mathrm{neg}}^{(j)}\|_2 \|T_{\mathrm{obj}}^{\mathrm{neg},(j)}\|_2}
\end{equation}

The cross-modal alignment loss $\mathcal{L}_{\mathrm{align}}$ constrains the alignment between the negation feature $T_{\mathrm{neg}}$ and the image feature $I$, preventing semantic drift:
\begin{equation}
    \mathcal{L}_{\mathrm{align}} = \frac{1}{B} \sum_{j=1}^{B} \left| \frac{T_{\mathrm{neg}}^{(j)} \cdot I^{(j)}}{\|T_{\mathrm{neg}}^{(j)}\|_2 \|I^{(j)}\|_2} - \frac{T_{\mathrm{obj}}^{\mathrm{neg},(j)} \cdot I^{(j)}}{\|T_{\mathrm{obj}}^{\mathrm{neg},(j)}\|_2 \|I^{(j)}\|_2} \right|
\end{equation}

Additionally, we implement a dynamic loss balancing strategy, starting with a higher emphasis on semantic similarity ($\delta = 0.5$) and gradually shifting weight toward the cross-modal constraint ($\delta = 1.0$) to stabilize negation representation learning before enforcing image alignment.

\noindent\textbf{Stage 2: Independent Training of the Frame.}
In this stage, the ground-truth negation feature $T_{obj}^{neg} \in \mathbb{R}^D$ is directly used as input to the Frame module, serving as the predicted negation feature from the Lens. Training is based on a generalized InfoNCE loss:

\begin{equation}
\mathcal{L}_{\text{gen}}(q, k^+, \mathcal{N}) = -\log \frac{e^{S(q, k^+)}}{e^{S(q, k^+)} + \sum_{k^- \in \mathcal{N}} e^{S(q, k^-)}}
\label{eq:generalized_infonce}
\end{equation}

where $q$ denotes the query feature, $k^+$ the corresponding positive key, and $\mathcal{N}$ the set of negative keys $k^-$. The similarity function $S(\cdot, \cdot)$ is defined as in Eq.~\eqref{eq:final}.

Depending on the availability of hard negatives, the loss is applied in two ways. When explicit hard negatives are available (\textit{e.g.}, from the CCNeg dataset), training is performed using constructed triplets \((q, k^+, k^-)\). Otherwise, in-batch contrastive learning is adopted, treating all other batch samples as implicit negatives.

\noindent\textbf{Stage 3: Joint Training of Lens and Frame.}
After training the Lens and Frame separately, we further design a joint training process to optimize their collaborative effect. Specifically, based on the training strategy of Stage 2, the ground-truth negated object feature $T_{obj}^{neg} \in \mathbb{R}^D$ in the input of the Frame is replaced with the output $T_{neg}$ of the Lens module, while the rest of the settings remain the same as in Stage 2.

\section{Comparative Experiments}
To validate our method, we conduct systematic comparisons with state-of-the-art baselines, evaluate the retention of CLIP's zero-shot capabilities, and perform ablations to assess the contributions of individual components.

\subsection{Comparative Analysis}

Existing work typically fine-tunes CLIP’s text encoder without architectural changes. Given their structural homogeneity—differing primarily in training data—we focus on two fully open-source, paradigm-representative baselines: NegCLIP~\cite{NegClip} and CoN-CLIP~\cite{conclip}. Comparative results are shown in Table~\ref{tab:compare_tbl}.

Trained on the CC-Neg dataset~\cite{conclip} (188K images, 376K captions), CLIPGlasses achieves 96.56\% accuracy on the in-domain CC-Neg-val set. While slightly below CoN-CLIP’s 99.70\%, this reflects a deliberate design choice to prioritize generalization over overfitting. The benefit becomes evident on the cross-domain Neg-COCO-MCQ benchmark~\cite{alhamoud2025vision}, where CLIPGlasses surpasses CoN-CLIP by 8.81 percentage points (34.51\% vs. 25.70\%).

Under low-resource conditions using the Neg-COCO-R subset (5K images, 22K captions; \citealp{alhamoud2025vision}), this advantage becomes more pronounced: CLIPGlasses outperforms CoN-CLIP by 27.45 points on CC-Neg-val (93.36\% vs. 65.91\%) and by 5.29 points on Neg-COCO-MCQ (35.90\% vs. 30.61\%).

These results highlight two core limitations of fine-tuning-based approaches: reliance on large-scale data, as seen in CoN-CLIP’s drop under constraints, and the absence of explicit negation modeling, as reflected in NegCLIP’s persistently low performance. In contrast, our non-invasive architecture maintains consistent performance across settings, validating its strength in capturing transferable, semantically grounded negation.

\subsection{Inherent Ability Retention Analysis}
A critical concern in enhancing negation understanding is whether such improvements come at the cost of a model’s inherent strengths. To assess this, we compare zero-shot performance on standard non-negation benchmarks (ImageNet~\cite{deng2009imagenet} and Caltech101~\cite{li2022caltech}) after training on negation datasets.

As shown in Table~\ref{tab:retention_tbl}, \textsc{CLIPGlasses} retains near-original performance, matching or even surpassing the vanilla CLIP on both datasets. In contrast, the CoN-CLIP shows notable degradation, particularly on ImageNet. These results confirm that our non-invasive design preserves CLIP's general visual-language alignment ability while enhancing negation understanding. 

\begin{table}[t]
    \centering
    \setlength{\tabcolsep}{5pt}
    \resizebox{0.96\columnwidth}{!}{
    \begin{tabular}{lccc}
        \toprule
        Method & Training Set & ImageNet & caltech101 
        \\
        \midrule
        Vanilla CLIP & - & \textbf{53.87} & \underline{90.96}
        \\
        CoN-CLIP & CC-Neg & 50.98 & 88.91 
        \\
        \textsc{CLIPGlasses} & Neg-COCO-R &  \underline{53.51} & 90.54 
        \\
        \textsc{CLIPGlasses} & CC-Neg & 53.28 & \textbf{90.97 }
        \\
        \bottomrule
    \end{tabular}
    }
    \caption{Zero-shot performance on standard non-negation benchmarks for models trained on negation datasets.}
    \label{tab:retention_tbl}
\end{table}

\begin{table}[t]
    \centering
    \small
    \setlength{\tabcolsep}{5pt}
    \resizebox{0.99\columnwidth}{!}{
        \begin{tabular}{lcccc}
            \toprule
            \textbf{Ablation Setting} & Acc & $\Delta$Acc & FAR & $\Delta$FAR \\
            \midrule
            \multicolumn{5}{l}{\textit{-------------------------------------Lens Module---------------------------------}} \\
            w/o Syntax Stream & 94.09\% & -2.47\% & 4.4927 & -3.3326 \\
            w/o Semantic Stream & 94.93\% & -1.63\% & 6.0925 & -1.7328 \\
            w/o Residual Gating & 68.93\% & -27.63\% & 0.7875 & -7.0378 \\
            \midrule
            \multicolumn{5}{l}{\textit{-------------------------------------Frame Module--------------------------------}} \\
            w/o Cross-modal & 93.79\% & -2.77\% & 7.3051 & -0.5202 \\
            w/o Repulsion weight & 63.74\% & -32.82\% & 0.5684 & -7.2569 \\
            \midrule
            Full Model & \textbf{96.56\%} & -- & \textbf{7.8253} & -- \\
            \bottomrule
        \end{tabular}
    }
    \caption{Ablation on CC-Neg evaluating the \textit{Lens} and \textit{Frame} modules. ``Acc'' and ``FAR'' refer to Accuracy and false alignment rate respectively, $\Delta$ denotes the performance drop relative to the full model.}
    \label{tab:ablation_tbl}
\end{table}

\subsection{Ablation Analysis}
To assess the individual contributions of key components, we conduct hierarchical ablation experiments on the CC-Neg dataset. Using the full model as our reference standard, we evaluate two key metrics: classification accuracy (Accuracy) and the average confidence margin between positive and negative pairs, termed false alignment rate (FAR), which is defined as:
\begin{eqnarray}\label{}
    \text{FAR} = \frac{1}{N} \sum_{i = 1}^{N} \left| s^+_i - s^-_i \right|
\end{eqnarray}
where \(s^+_i\) and \(s^-_i\) denote the similarity scores between the image and the positive and negative captions in the \(i\)-th sample, and \(N\) is the total number of samples. Higher FAR indicates stronger discriminative ability between affirmations and negations. The results of the ablation experiments are shown in Table~\ref{tab:ablation_tbl}.

\noindent\textbf{Effect of Syntactic Stream.}
Removing the syntactic stream results in a 2.47\% decrease in accuracy and a more substantial 3.33-point decline in FAR, demonstrating its critical role in enhancing the model's structural sensitivity to negation.
In particular, the reduction in FAR indicates diminished confidence in negation detection, underscoring the essential role of syntactic grounding in achieving reliable alignment.

\noindent\textbf{Effect of Semantic Stream.}
Excluding the semantic stream results in a 1.63\% accuracy drop and a 1.73-point reduction in FAR. These results indicate that while syntactic patterns effectively locate negation cues, accurate interpretation requires access to global semantic context. The semantic stream, derived from the final CLIP layer, enables the model to identify the target of negation by capturing sentence-level meaning. Without this semantic grounding, the model exhibits reduced performance in disambiguating negation, particularly when polarity lacks explicit syntactic markers.

\noindent\textbf{Effect of Residual Gating.}
Disabling the residual gating leads to a substantial 27.63\% drop in accuracy and a 7.04-point reduction in FAR, demonstrating severe degradation in both classification and discriminative confidence. This validates the critical importance of balancing syntactic attention with the original semantic representation. Without residual gating, the model becomes susceptible to overfitting on structurally salient yet semantically irrelevant patterns, particularly in linguistically ambiguous contexts. The gating mechanism enables the model to preserve core sentence meaning when structural alignment is uncertain, thereby enhancing the robustness and stability of negation reasoning.

\noindent\textbf{Effect of Cross-modal Context.}
Disabling cross-modal interaction results in a 2.77\% accuracy decline and a 0.52-point reduction in FAR. Under this configuration, the repulsion weight $\lambda$ is estimated solely from textual features, thereby limiting the model’s ability to adjust repulsion strength based on visual content--a critical component for context-aware negation understanding.

\noindent\textbf{Effect of Dynamic Repulsion Weight.}
We observe substantial performance degradation upon removing this module (32.82\% accuracy decrease) underscores its critical importance.
This ablation result motivates a deeper investigation into the underlying mechanisms driving its contribution.
We hypothesize that the module's primary function is to dynamically calibrate repulsion strength according to the linguistic intensity of negation. To test this hypothesis, we evaluate the predicted \(\lambda\) values across a controlled spectrum of negation intensities. Using Qwen2.5-72B~\cite{qwen2.5}, we generate four recaptioned variants for 500 randomly sampled CC-Neg examples, each corresponding to four distinct levels of negation strength: strong (``\emph{no}''), moderate (``\emph{not any}''), weak (``\emph{appears to be absent}''), and weakest (``\emph{may not be}'').

\begin{figure}[t]
    \centering
    \includegraphics[width=\columnwidth]{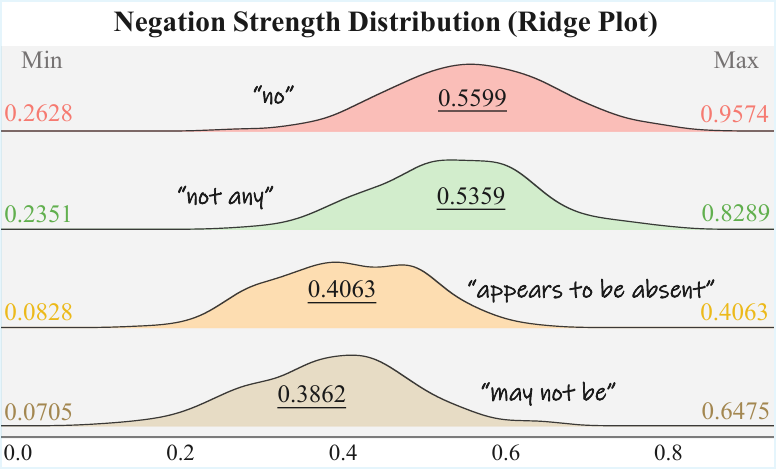}
    \caption{Distribution of predicted repulsion weight \(\lambda\) under varying negation strengths. Stronger negations (\textit{e.g.}, ``\emph{no}'') yield higher \(\lambda\), confirming the model's ability to adaptively modulate semantic repulsion.}
    \label{fig:lambda_dist}
\end{figure}

As shown in Figure~\ref{fig:lambda_dist}, the predicted \(\lambda\) distributions exhibit a consistent decreasing trend aligned with the reduction in negation strength. These results demonstrate that the module adaptively modulates repulsion intensity in response to linguistic cues.

\section{Conclusion}
In this paper, we present \textsc{CLIPGlasses}, a novel framework that addresses CLIP’s limitations in negation understanding. Our non-intrusive design introduces cognitively inspired modules: a \textit{Lens} for disentangling negated semantics and a \textit{Frame} for modeling context-aware repulsion, together enabling negation-sensitive similarity computation without modifying CLIP’s pretrained parameters. Experiments show that our method achieves competitive in-domain accuracy, state-of-the-art cross-domain generalization and low-resource robustness, all while preserving CLIP’s native zero-shot capabilities. Nevertheless, a shared limitation across current methods remains in handling non-visual negations (\textit{e.g.}, ``\emph{not authentic}''). Future work will explore integrating commonsense knowledge to address such cases.

\section{Acknowledgements}
This work was supported by the National Natural Science Foundation of China (General Program, No. 62377024, 2024–2027).


\end{document}